\documentclass[10pt,twocolumn,letterpaper]{article}

\usepackage{cvpr}              %
\usepackage{comment}
\usepackage{times}
\usepackage{epsfig}
\usepackage{graphicx}
\usepackage{amsmath}
\usepackage{amssymb, xcolor}
\usepackage{tabularx}
\usepackage{booktabs}
\usepackage{arydshln}
\usepackage{subcaption}
\usepackage{tikz}
\usepackage{pgfplots}
\usepackage{multirow}
\usepackage{multicol}
\usepackage{enumitem}
\usepackage{hhline}
\usepackage{boldline,microtype}
\usepackage{bm}
\usepackage{enumitem}
\usepackage{makecell}
\usepackage{hhline}
\usepackage{xcolor}
\usepgfplotslibrary{fillbetween}

\DeclareUnicodeCharacter{2061}{}

\newcommand{\jj}{\mathbf{j}}
\newcommand{\kk}{\mathbf{k}}
\newcommand{\ii}{\mathbf{i}}

\definecolor{col1}{HTML}{FE6100}
\definecolor{col2}{HTML}{DC267F}
\definecolor{col3}{HTML}{785EF0}
\definecolor{col4}{HTML}{648FFF}

\definecolor{cvprblue}{rgb}{0.21,0.49,0.74}
\usepackage[pagebackref,breaklinks,colorlinks,citecolor=cvprblue]{hyperref}

\newcommand{\mname}{DiffAssemble}
\newcommand{\bmname}{\textbf{\mname}}
\newcommand{\dropout}{Sparse Attention Mechanism}
\usepackage{pifont}

\usepackage{array}
\usepackage{pifont}
\newcolumntype{P}[1]{>{\centering\arraybackslash}p{#1}}
\newcolumntype{H}{>{\setbox0=\hbox\bgroup}c<{\egroup}@{}}

\def\paperID{3975} %
\def\confName{CVPR}
\def\confYear{2024}

\title{DiffAssemble: A Unified Graph-Diffusion Model for 2D and 3D Reassembly}

\author{Gianluca Scarpellini$^*$\quad Stefano Fiorini$^\textbf{*}$\quad Francesco Giuliari$^\textbf{*}$\quad Pietro Morerio\quad Alessio Del Bue\\
Pattern Analysis and Computer Vision (PAVIS) \\
Istituto Italiano di Tecnologia (IIT) \\
$*$ Equal contributions
}

\begin{document}
\maketitle
\begin{abstract}
Reassembly tasks play a fundamental role in many fields and multiple approaches exist to solve specific reassembly problems. In this context, we posit that a general unified model can effectively address them all, irrespective of the input data type (images, 3D, etc.). 
We introduce \bmname, a Graph Neural Network (GNN)-based architecture that learns to solve reassembly tasks using a diffusion model formulation.
Our method treats the elements of a set, whether pieces of 2D patch or 3D object fragments, as nodes of a spatial graph. Training is performed by introducing noise into the position and rotation of the elements and iteratively denoising them to reconstruct the coherent initial pose.
\bmname\ achieves state-of-the-art (SOTA) results in most 2D and 3D reassembly tasks and is the first learning-based approach that solves 2D puzzles for both rotation and translation. Furthermore, we highlight its remarkable reduction in run-time, performing 11 times faster than the quickest optimization-based method for puzzle solving. Code available at \url{https://github.com/IIT-PAVIS/DiffAssemble}.

\end{abstract}

\section{Introduction}\label{sec:intro}

\emph{Spatial Intelligence} is the ability to perceive the visual-spatial world accurately and to perform transformations upon the perceived space~\cite{gardner2011frames}. %
This skill is commonly assessed with reassembly tasks, which involve arranging and connecting individual components to form a coherent and functional entity. Examples of such tasks include solving 2D jigsaw puzzles or assembling 3D objects with LEGO blocks.
Since the proposal of the first puzzle solver~\cite{freeman1964apictorial}, \emph{Spatial Intelligence} has
challenged the Machine Learning (ML) community with its intrinsic combinatorial complexity and its numerous applications, such as genomics~\cite{marande2007mitochondrial}, assistive technologies~\cite{yurteri2022obstacle}, fresco reconstruction~\cite{brown2008system, villegas2021matchmakernet} and molecular docking \cite{corso2023diffdock}.

\begin{figure}[t]
\centering
\includegraphics[width=\linewidth]{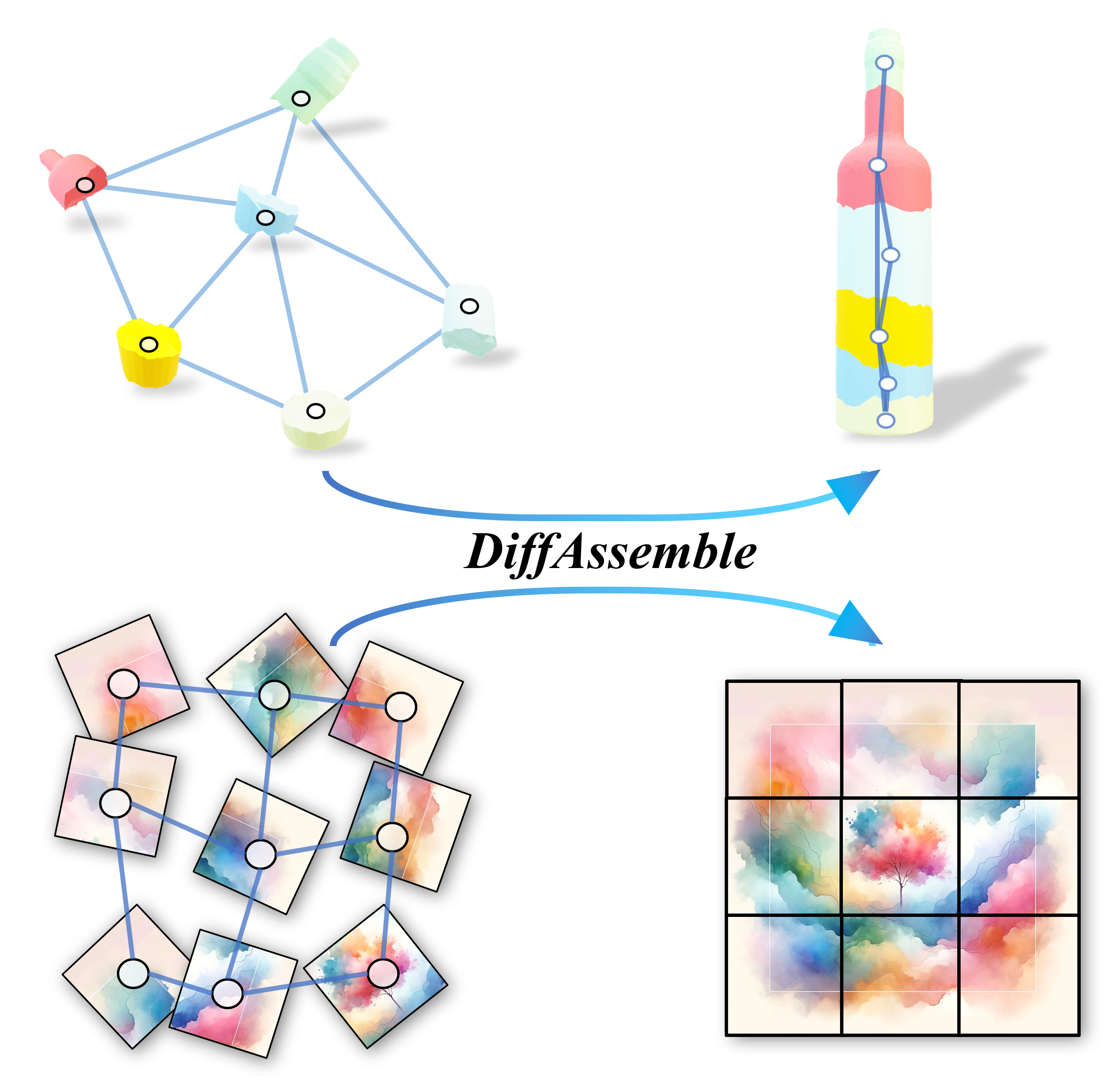}

\caption{\label{fig:fig1} We propose \bmname\ as a unified approach to deal with reassembly tasks in two and three dimensions. \mname\ processes the elements to reassemble as a graph and infer their correct position and orientation in 2D and 3D space.}
\vspace{-0.7cm}
\end{figure}

Reassembling a set involves placing each element in its correct position and orientation to form a coherent structure, that being a 2D jigsaw puzzle or a 3D object, as in Figure~\ref{fig:fig1}. Despite the similarities between the tasks, the literature addresses %
reassembly tasks in different dimensions separately.

In the 2D dimension, the most common reassembly problem is related to the resolution of puzzles, particularly those with pieces that are translated and rotated and have a regular shape, i.e., square pieces of the same dimension. Due to the regularity of the pieces, the problem can be treated as a permutation problem and solved via optimization-based approaches~\cite{gallagher2012jigsaw,yu2015solving,huroyan2020solving}. These solutions are effective but lack robustness, showing a massive drop in performance when dealing with non-standard scenarios, such as eroded or missing pieces~\cite{talon2022ganzzle}. On the other hand, recent learning-based solutions are robust to distortion in the visual aspect of the pieces by working in the feature space but cannot handle rotations and perform worse than greedy approaches in the standard scenario. 

Regarding the 3D reassembly task, since the 3D pieces are not regular, it can not be solved as a permutation problem but has to be solved in the continuous domain, making it a much more challenging task where optimization-based solutions cannot be applied. As a part of the ongoing efforts to address fractured 3D object reassembly, Sellan \etal recently introduced the \textit{Breaking Bad} dataset~\cite{sellan2022breaking} that includes fragments of thousands of 3D objects, and it is commonly adopted as a benchmark for 3D object reassembly solutions. 
Despite the interest of the machine learning community, the results achieved in 3D reassembly tasks have yet to reach the same level of performance as their 2D counterparts due to the increased complexity of the task.

We argue that 2D jigsaws and 3D objects are two aspects of the same problem, namely reassembly. All these tasks share some properties and, potentially, common solutions. Nonetheless, methods that tackle only one of these tasks are too narrow to generalize to the others. A unique approach that tackles all reassembly tasks at once may benefit from their shared characteristics.

This work introduces \bmname, a general framework for solving reassembly tasks using graph representations and a diffusion model formulation.
In contrast to prior learning-based approaches for reassembly tasks, which typically tackle the problem in a single step, our approach uses a multi-step solution strategy leveraging Diffusion Probabilistic Models (DPMs) to guide the process.
We represent the elements to be reassembled using a %
graph formulation, allowing us to work with an arbitrary number of pieces.
Each piece is modeled as a node that contains the piece's visual appearance, extracted with an equivariant encoder, and the piece's position and orientation. By mapping the appearance to a latent space, we can remove the separation that exists between 2D and 3D tasks and propose a unique solution. 

We structure the learning problem using the Diffusion Probabilistic Models (DPM) formulation.
We iteratively add Gaussian Noise to each piece's starting position and orientation until they are randomly placed in the Euclidean space. We then train an Attention-based Graph Neural Network~\cite{shi-graphtransformer} to reverse this noising process and retrieve the pieces's original pose from a random starting position and orientation.
By adopting a sparsifying mechanism~\cite{shirzad2023exphormer} on the graph, we run \mname\ on graphs with up to 900 nodes with minimal loss in accuracy while greatly reducing the memory requirement.

Our solution achieves state-of-the-art performance in most 2D and 3D tasks, showcasing that these tasks share common characteristics and can thus be solved through a unified approach.
In \textit{2D,} compared to optimization-based solutions, our solution is more robust to missing pieces and much faster, i.e., 5 seconds to rearrange 900 pieces compared to 55 seconds for the fastest optimization approach.
In \textit{3D,} our method achieves state-of-the-art results in both rotation and translation accuracy without sacrificing one for the other, as is the case for previous learning-based solutions.

\paragraph{Main Contributions and Novelty of the Work:}
\begin{itemize}

    \item We propose \bmname, a unified learning-based solution using diffusion models and graphs neural networks for reassembly tasks that achieve SOTA results in most 2D and 3D without distinguishing between the two.
    
    \item We show that reassembly tasks in 2D and 3D share several key properties and that model choices such as the use of different losses, different diffusion chains, and equivariant features.

    \item To the best of our knowledge, \mname{} is the first learning-based solution that can handle rotations and translations for 2D visual puzzles.

\end{itemize}

\section{Related Works}

\begin{figure*}[th]
\centering
\includegraphics[width=\linewidth,height=8.8cm, trim={1.1cm 0.8cm 0.2cm 0.3cm},clip]{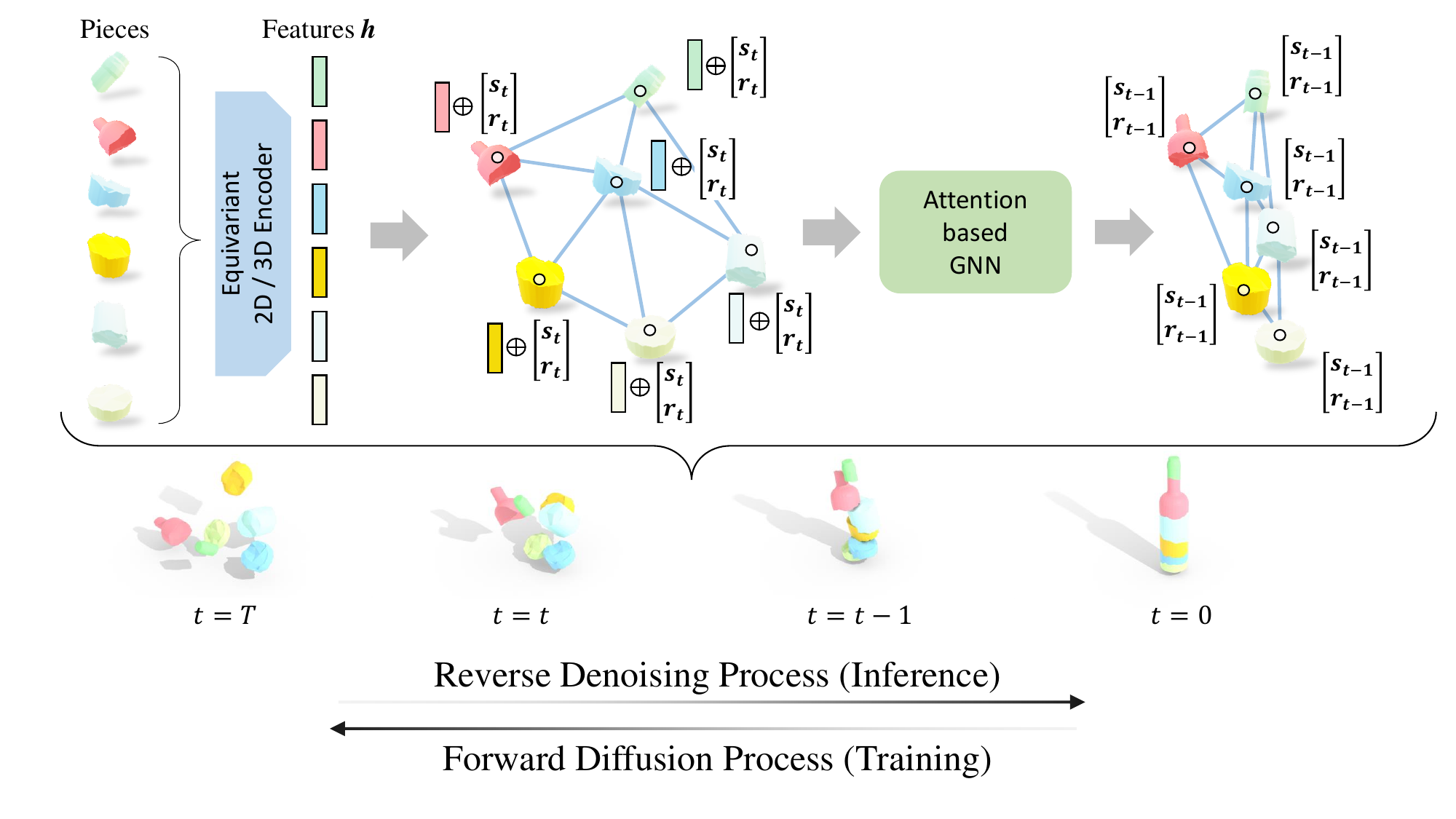}

\caption{Framework of our proposed \mname\ for reassembly tasks, here is shown for the 3D task. Following the Diffusion Probabilistic Models formulations, we model a Markov chain where we inject noise into the pieces' position and orientation. At timestep $t = 0$, the pieces are in their correct position, and at timestep $t = T$, they are in a random position with random orientation.
At each timestep $t$, our attention-based GNN takes as input a graph where each node contains an equivariant feature that describes a particular piece and its position and orientation. The network then predicts a less noisy version of the piece's position and orientation. }
\label{fig:network_arc}
\vspace{-.4cm}
\end{figure*}

\label{sec:SOTA}

In this section, we revise the main literature %
for reassembly tasks in 2D and 3D.
We complement the discussion %
with a brief overview of %
recent advancements in diffusion models and graph neural networks.

\paragraph{Reassembly Tasks.} 
Reassembly tasks captivate the attention of the research community as a benchmark for investigating the effectiveness of solutions that employ a reasoning process in the spatial domain. %
Here, we present the relevant works for the most common reassembly task in 2D and 3D: jigsaw puzzles and fracture object reassembly.

\emph{2D jigsaw puzzles.} 
Puzzles are used to investigate the intricacies of image ordering with inherent combinatorial complexity~\cite{choCVPR10probjigsaw}. 
Among the most successful strategies are those rooted in optimization and greedy approaches that rely on hand-crafted features~\cite{gallagher2012jigsaw,yu2015solving,huroyan2020solving}.
More recently, there has been a shift towards employing learning-based methods to solve puzzles with only shifted pieces~\cite{talon2022ganzzle, paumard2020deepzzle, Li2021JigsawGANAL, giuliari2023positional}. These approaches demonstrate greater resilience when handling inputs with distortions, though they %
perform worse compared to optimization methods in standard scenarios. Moreover, these methods do not handle rotated pieces. %

\emph{3D fractured objects.} Fractured object reassembly in 3D is extensively explored in the literature~\cite{chen2022neural, funkhouser2011learning} and has applications in numerous fields, such as fresco reconstruction~\cite{brown2008system}, and furniture assembly~\cite{lee2021ikea}. A recent effort in solving the problem was introducing the \textit{Breaking Bad} dataset~\cite{sellan2022breaking}. In \textit{Breaking Bad}, the challenge involves reconstructing a broken object from multiple fragments. Those fragments are not labeled with any semantic information, %
as in many real-world applications~\cite{brown2008system}. Previous research efforts focus on predicting 6-degree-of-freedom poses for input parts (such as chair backs, legs, and bars)~\cite{zhan2020generative} and assembling 3D shapes from images of the complete object~\cite{li2020learning}. These prior investigations heavily lean on the semantic details of object parts, overlooking essential geometric cues. Neural Shape Matching (NSM)~\cite{chen2022neural} addressed the two-part mating problem by emphasizing shape geometries over semantic information. \textit{SE(3)-Equiv}~\cite{wu2023leveraging} tackles the problem with specific design choices that go beyond object reassembly, e.g., adversarial and reconstruction losses.

Unlike the previous works, which focus on just one aspect of the problem, we propose a unified solution for reassembly tasks. %
Furthermore, to the best of our knowledge, we are the first to propose a learning-based approach for puzzles with translated and rotated pieces.

\paragraph{Diffusion Probabilistic Models.}
DPMs are generative models that have shown remarkable results in recent years. These models approach the generation process through a bidirectional iterative chain. In one direction, they transform data into a Gaussian distribution by incrementally adding Gaussian noise. They are then trained to reverse this process, generating new samples from the initial distribution starting from random noise~\cite{ho2020denoising}. DPMs demonstrate remarkable versatility across a range of tasks %
applications, including image synthesis~\cite{dhariwal2021diffusion}, semantic segmentation~\cite{baranchuk2021label}, %
and generation~\cite{luo2021diffusion}. Their applicability extends to spatial data processing in both 2D and 3D contexts, where they have been effectively employed in object detection~\cite{chen2023diffusiondet}, scene generation~\cite{huang2023diffusion}, and 3D protein modeling~\cite{yim2023se}. %

Our work proposes the use of Diffusion Models for reassembly tasks. We introduce a unified model capable of operating effectively in both 2D and 3D space, aiming to retrieve the original position and orientation of the constituent pieces accurately.

\paragraph{Graph Neural Networks.}
Graph Neural Networks (GNNs) underwent significant advancements in recent years with new models like GCN~\cite{kipf2016semi}, GraphSage~\cite{liu2020graphsage}, and SigMaNet~\cite{fiorini2023sigmanet}. These advancements have continued with new architectures~\cite{velivckovic2018graph, zhang2020graph} that incorporate attention mechanisms that weigh the importance of nodes during message passing.
The use of Graphs and GNNs has seen widespread adoption for spatial applications, as they are able to describe an arbitrary number of elements and their relation to each other. Common applications include scene graph generation~\cite{Rosinol20rss-dynamicSceneGraphs}, 3D scene generation~\cite{graph2scene2021}, object localisation~\cite{giuliari2022spatial}, relative pose estimation~\cite{taiana2022posernet}, and robot navigation~\cite{Ravichandran2022Scenenavigation}.
Scalability is a common issue affecting GNNs and Exphormer~\cite{shirzad2023exphormer} represents a significant stride in scalable graph transformer architectures, utilizing a sparse attention mechanism that leverages virtual global nodes and expander graphs~\cite{deac2022expander}.

We represent the puzzle as a graph and process it by using an Attention-based GNN~\cite{shi2020masked}. We adopt the Exphormer~\cite{shirzad2023exphormer} to enhance \mname's capability in handling the computational demands of reassembly tasks. %

\section{Our Method} \label{sec:method}

We reassemble a set of elements by predicting the translation and rotation, i.e. the pose of a piece, 
with the objective to arrange %
the elements in a coherent structure, such as a non-broken 3D object or a solved 2D puzzle. 
Figure~\ref{fig:network_arc} summarizes our approach. 
We represent the set of pieces to reassemble as nodes in a complete graph, %
with each node having a modality-specific feature encoder. We adapt the DPM formulation~\cite{ho2020denoising} by introducing time-dependent noise into each element's translation and rotation. Noise injection resembles shuffling the set of elements, e.g., randomly distributing puzzles pieces or 3D object fragments in the 2D or 3D Euclidean space.
During training, we process the noisy input graph via an Attention-based GNN to restore the initial translation and rotation of all elements. At inference, we initialize each element's starting positions and rotation from pure noise, and iteratively denoise the graph, reassembling the coherent structure in the process. %
Section \ref{sec:input} introduces our graph-based formulation. Section \ref{sec:features} discusses the input's feature representation. Section \ref{sec:dpm} presents our diffusion-based approach, and Section \ref{sec:arch} defines our attention-based architecture and the sparsity mechanism.

\subsection{Graph Formulation}
\label{sec:input}
We assign each of the $M$ pieces $m$ to a node $v^m$, defining a set of vertices $V = \{v^m\}_{m \in [1, \dots, M]}$. Since we do not want to introduce a priori relationship between the pieces, we connect all the nodes together.
This defines a complete graph $G = (V, E)$, where %
$E$ is the set of edges.
We define the feature of each node $v^m$ %
by concatenating the following vectors:
\begin{itemize}
    \item \textbf{Features vector} ${\textbf{h}}^m \in \mathbb{R}^{d}$, where $d$ is the dimension of the feature generated by a equivariant encoder. \mname\ is agnostic to the adopted feature backbone. %
    \item \textbf{Translation vector} ${\textbf{s}}^m \in~\mathbb{R}^n$, where $n$ represents the dimensionality of the \textit{continuous} Euclidean space in which the reassembly task is conducted. 
    \item \textbf{Rotation matrix} ${{R}}^m \in SO(n)$, where $SO(n)$ is the Special Orthogonal Group in $n$ dimensions. We also define a function $f_r(\textbf{r}^m) = R^m$ that maps any vector rotation representation $\textbf{r}^m$ to $R^m$.
\end{itemize}
The advantage of using this graph formulation lies in its ability to be flexible with respect to the cardinality of $V$. For this reason, \mname{} is able to work simultaneously with puzzles of various sizes rather than being limited to handling only one size at a time.

\subsection{Feature Representation}
\label{sec:features}

A fundamental aspect of our architecture lies in its capability to operate with element features $\textbf{h}^m$, which can be extracted by pre-trained encoders. Features play a central role in solving reassembly tasks, as they provide the network with inductive biases. %
This intuition is particularly relevant for complex tasks involving translation and rotation.

To extract the features, we first translate the piece so that its center lies on the origin and then use a rotation-equivariant encoder to map the visual and shape information into the latent space.
Rotation-equivariant features undergo the same rotation that is applied in the original input space and are thus the best candidates to enable the neural network associating a specific rotation $R^m$ (in the input space) to the features map $\textbf{h}^m$. More details on group equivariance are given in the Supplementary Material.

\subsection{Diffusion Models for Reassembly Tasks}
\label{sec:dpm}
We adopt Diffusion Probabilistic Models as defined in DDIM~\cite{song2020denoising} to solve the reassembly tasks.
We define a compact representation of the initial translation $\textbf{s}^m_0$ and rotation $\textbf{r}_0^m$ of piece $m$ as a concatenated vector $\textbf{x}_0^m = [\textbf{s}_0^{m^{\top}},\textbf{r}_0^{m^{\top}}]^\top$. %

At training time, we iteratively add noise sampled from a Gaussian Distribution $\mathcal{N}(0, I)$ to their poses (Forward Process).
Following that, we train %
\mname{} to reverse this noising process (Reverse Process) and to obtain the initial poses $X_0 = \{\textbf{x}_0^m\}_{m \in [1,\cdots, M]}$.

\paragraph{Forward Process.}
We define the forward process as a fixed Markov chain that adds noise following 
a Gaussian distribution to each input $\textbf{x}_0^m$ to obtain a noisy version, $\textbf{x}_t^m$, at timestep $t$.
Following~\cite{ho2020denoising}, we adopt the variance $\beta_t$ according to a linear scheduler and define $q(\textbf{x}_t^m | \textbf{x}_0^m)$ as:
\begin{equation}\label{eq:forward}
    q(\textbf{x}_t^m | \textbf{x}_0^m) = \mathcal{N}(\textbf{x}_t^m; \sqrt{\overline \alpha_t} \textbf{x}_0^m, (1 - \overline \alpha_t) \textbf{I}), 
\end{equation}
where $\overline \alpha_t = \prod_{c=1}^{t} (1 - \beta_c)$ and $\textbf{I}$ is the indentity matrix.

\paragraph{Reverse Process.}
The reverse process iteratively retrieves the initial poses for the set of elements ${\hat{X}}_{t-1}$ given current (noisy) poses ${X}_t = \{\textbf{x}^m_t\}_{m \in [1,\cdots, M]}$ and the features ${H} = \{\textbf{h}^m\}_{m \in [1,\cdots, M]}$. ${\hat{X}}_{t-1}$ is computed as:

\begin{equation}\label{eq:backward}
{\hat{X}}_{t-1} = \frac{1}{\sqrt{\alpha_t}}\left(  {X}_t - \frac{1-\alpha_t}{\sqrt{1 - \overline{\alpha}_t}}\epsilon_\theta({X}_t,{H},{t}) \right),
\end{equation}

\noindent where $\alpha_t = 1 - \beta_t$, and $\epsilon_\theta ({X}_t, H, {t})$ is the estimated noise output by \mname{} that has to be removed from ${\hat{X}}_t$ at timestep $t$ to recover ${\hat{X}}_{t-1}$. %

\paragraph{Losses.} %
Following a standard practice in Diffusion Models~\cite{ho2020denoising}, we train \mname{} to predict $\hat{X}_0$ instead of $\hat{X}_{t-1}$.
We introduce two loss functions to reconstruct the intial pose of each piece. %

\textbf{Translation Loss.} This loss computes the average difference between the ground truth translation vectors and the predicted ones $\hat{\mathbf{s}}^m_0$: %
    \begin{equation*}
        \mathcal{L}_{tr} = \frac{1}{M}\sum_{m=1}^{M}|| \mathbf{s}_0^m - \hat{\mathbf{s}}^m_0 ||_2^2,
    \end{equation*}
    where $|| \cdot ||^2_2$ is the squared L2 norm. %

\textbf{Rotation Loss.} This loss measures the average difference between the ground truth rotation matrices and the predicted ones $f_r{(\hat{\textbf{r}}}^{m}_0)$: %
    \begin{equation*}
    \mathcal{L}_{rt} = \frac{1}{M} \sum_{m = 1}^{M} ||f_r(\textbf{r}_0^{m})^\top f_r{(\hat{\textbf{r}}}^{m}_0) - \textbf{I} ||_2^2.
    \end{equation*}

\subsection{Architecture}\label{sec:arch}

We use an Attention-based GNN with $L-$layers of Unified Message Passing (UniMP)~\cite{shi-graphtransformer}. UniMP implements a multi-head attention mechanism over all nodes to scale the information gathered from neighboring nodes during message passing. Multi-head attention is well-suited for graph contexts where we lack prior knowledge of node relationships, i.e., we cannot define an adjacency matrix $A$. 

However, one of the main constraints associated with %
these attention-based %
architectures~\cite{velivckovic2018graph, zhang2020graph} lies in the inherent definition of a complete graph. This constraint poses a severe limitation for scaling on large graphs, i.e., dealing with a large number of elements.   %
To address this constraint, we employ exphormer~\cite{shirzad2023exphormer}, which relies on the expander graph~\cite{hoory2006expander} and virtual nodes to reduce memory requirements by cleverly pruning edges in the graph. In Section~\ref{sec:scaling}, we show how we use exphormer to scale to large graphs.

\section{Experimental Evaluation}\label{sec:experiment}
\mname\ tackles 3D objects reassembly and 2D jigsaw puzzles as two possible instantiations of a \textit{reassembly task}. 
We first validate \mname\ on 3D object reassembly (Section \ref{sec:3d}), showing through quantitive and quality results the benefits of our approach.
Section \ref{sec:2d} discusses the performance of our approach on 2D jigsaw puzzles in the standard scenario and when dealing with missing pieces.
We carry out an ablation study of %
\mname's design choices in Section \ref{sec:3d} and in the Supplementary Material. Finally, we tackle \mname's limitation on large puzzles in Section \ref{sec:scaling}, demonstrating that \mname\ efficiently reassemble up to 900 elements thanks to the sparsity mechanism.

Throughout this section, tables report the best results in \textbf{boldface} and the second-best \underline{underlined}. \newline

\subsection{3D Object Reassembly}
\label{sec:3d}
\begin{table*}[h]
\centering
\begin{tabularx}{\linewidth}{X | P{2cm} P{2cm} P{2cm}}
\hline
\multicolumn{1}{c|}{\multirow{2}{*}{
    \textsc{Method}}} & RMSE ($R$) $\downarrow$ & RMSE ($T$) $\downarrow$ & \multicolumn{1}{c}{PA $\uparrow$} \\ 
\multicolumn{1}{c|}{} & \multicolumn{1}{c}{degree} & \multicolumn{1}{c}{$\times 10^{-2}$} & \multicolumn{1}{c}{$\%$} \\ \hline
Global \cite{sellan2022breaking} & 81.6 & {15.2} & 17.5 \\
DGL \cite{sellan2022breaking} & 81.4 & \underline{14.9} & \underline{25.4} \\
LSTM \cite{sellan2022breaking} & 87.4  & 15.8 & 11.3 \\
NSM \cite{chen2022neural}$^\dagger$ & 83.3 & 15.3 & 10.6 \\

SE(3)-Equiv \cite{wu2023leveraging}  & \underline{77.9}  & 16.7 & 8.1 \\
\hdashline 
\bmname\ - No Diffusion Process & 83.6    & 17.1  & 3.1 \\ %
\bmname\ - No Equivariant Enc. & 81.7 & 17.0 & 18.3 \\

\bmname & \textbf{73.3}  & \textbf{14.8} & \textbf{27.5} \\

\end{tabularx}%
  \caption{Quantitative results of four learning-based shape reassembly baselines and \mname\ on the \texttt{everyday} object subset. $^\dagger$Modified version, suggested in~\cite{wu2023leveraging}, capable of handling multi-part assembly.}
  \label{table:bb_compare}
  \vspace{-.2cm}
\end{table*}
\begin{table*}[t]
\centering
\begin{tabularx}{\linewidth}{l|X | P{2cm} P{2cm} P{2cm}}
\hlineB{2}
\multicolumn{1}{c|}{\multirow{2}{*}{\textsc{Stage}}} & \multicolumn{1}{c|}{\multirow{2}{*}{\textsc{Changes}}}  & RMSE ($R$) $\downarrow$ & RMSE ($T$) $\downarrow$ & \multicolumn{1}{c}{PA $\uparrow$} \\ 
& & \multicolumn{1}{c}{degree} & \multicolumn{1}{c}{$\times 10^{-2}$} & \multicolumn{1}{c}{$\%$} \\ \hline
\multirow{2}{*}{Representation} 
& Non-Equivariant Enc. & 81.68 & 17.04 & 18.32 \\
& Invariant Enc. & 77.06  & 18.09  & 14.27 \\ \hdashline
 \multirow{4}{*}{Diff. Design} & 6 degree-of-freedom rotation & {75.60} & 18.80 & 18.50 \\
& w/o Chamfer Distance loss & \textbf{72.75}  & \textbf{14.78}  & \underline{24.10} \\ 
 & w/ Chamfer Distance loss & \underline{73.34}  & \underline{14.82} & \textbf{27.48} \\ 
 & No Diff. process & 83.60    & 17.12  & 3.10 \\
 \hdashline

\multirow{1}{*}{GNN} &Standard GCN~\cite{kipf2016semi} & 74.56 & 15.79  & 21.33 \\

\end{tabularx}%
 \caption{
Ablation for 3D object reassembly on the \texttt{everyday} object subset.}
  \label{tab:abl_3d}
  \vspace{-.5cm}
\end{table*}

First, we explore the application of \mname{} to the task of reassembling objects in 3D.

\paragraph{Dataset and Evaluation Setting.}

We test our methods on 3D object reassembly on Breaking Bad (BB)~\cite{sellan2022breaking}. It is composed of 3D meshes for 20 classes of everyday objects, such as bottles, plates, glasses. For each of these objects, there are multiple variants, where the object is broken into multiple parts by simulating fractures in the geometry. The dataset provides objects split into 2 to 100 pieces. As proposed in BB, we train and test using objects composed of 2 to 20 parts. 
All the objects' pieces are translated to the origin and randomly rotated. For each piece, we need to provide a pose that returns the fragment to its correct location in the object's canonical pose.
Following the evaluation pipeline in~\cite{sellan2022breaking}, we report the metrics in terms of Root Mean Squared Error in rotation RMSE (R), Root Mean Square Error in translation RMSE (T), and Part Accuracy (PA), which measures the percentage of parts whose Chamfer Distance to ground-truth is less than $0.01$~\cite{zhan2020generative}.
We compare with the three baselines proposed in BB: Global, DGL, LSTM. In addition, we compare with SE(3)-Equiv.~\cite{wu2023leveraging}, the current state of the art on BB, which integrates both equivariant and invariant features. Regarding our approach, we use the base model along with two variations: one without the diffusion process, predicting the translation and rotation of the pieces in a single step, and another version that omits the use of an equivariant encoder.

\begin{figure}[t]
\includegraphics[width=0.95\linewidth]{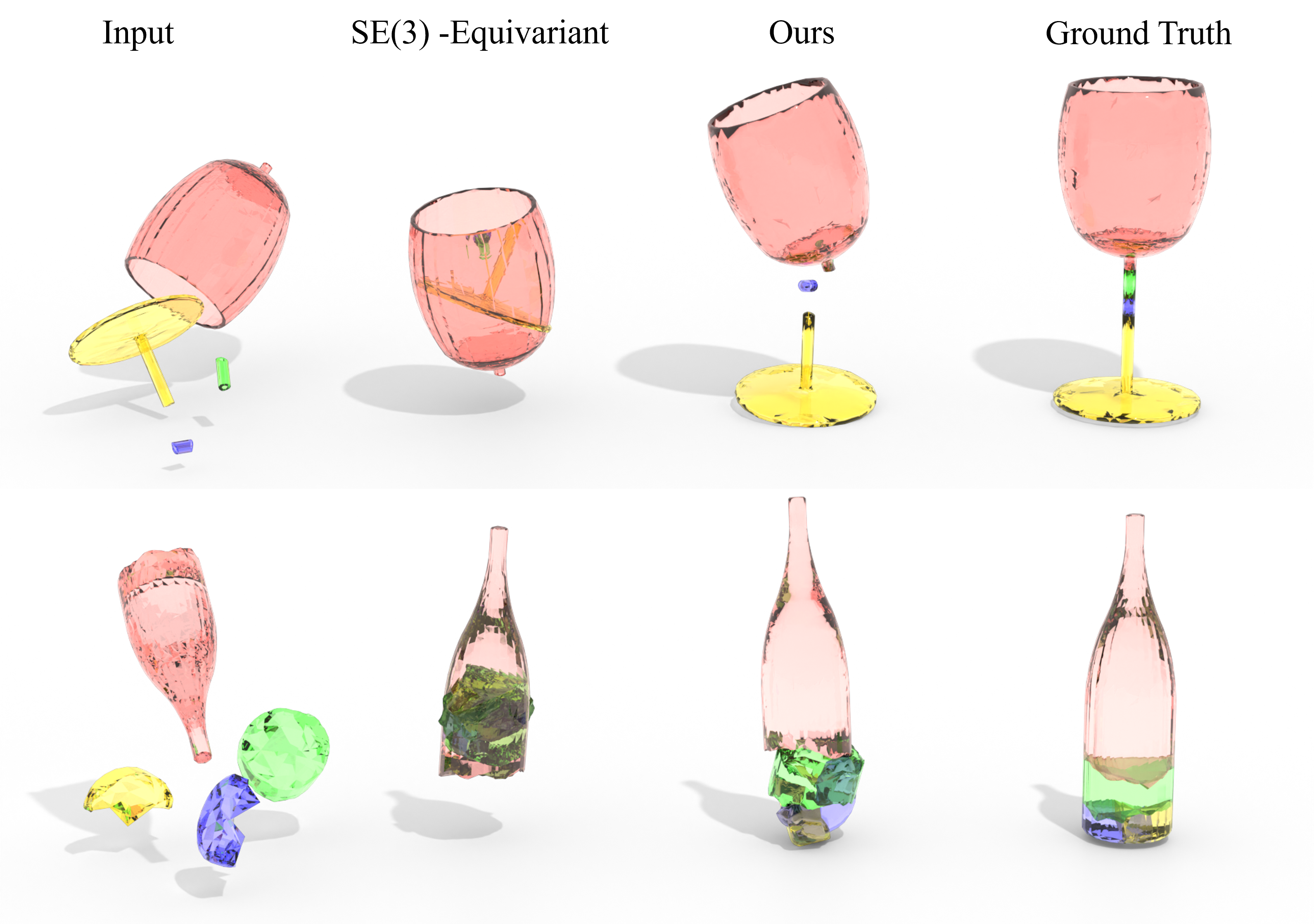}
\centering
\caption{\label{fig:qual_3d} Qualitative results on Breaking Bad, showing the reassembly results for a broken wine glass and a wine bottle. We compare the results against SE(3)-Equiv~\cite{wu2023leveraging}, which is the current SOTA method. All results are in the same reference frame, shifted horizontally so they do not overlap. We show the results with glass materials to better show overlapping pieces. }
\vspace{-0.5cm}
\end{figure}

\paragraph{Implementation Details.}

Each fragment $m$ of a 3D object is a point-cloud of $1,000$ points. 
For the 3D shape reassembly, we use VN-DGCNN~\cite{wu2023leveraging} as our feature extractor. This backbone takes as input the point cloud of each piece and returns both an equivariant and invariant representation.
We only consider the equivariant features to create the element features $\textbf{h}^m$.
For this task, we parameterize the rotation in 3D as a unit quaternion, 
    ${q}^m = q_0^m + \ii q_1^m + \jj q_2^m + \kk q_3^m ,$ 
where $q_0^m$, $q_1^m$, $q_2^m$ and $q_3^m$ are real numbers, $\ii$, $\jj$ and $\kk$ are mutually orthogonal basis vectors. Thus, we define the vector %
$
\textbf{r}^m = [q_0^m, q_1^m, q_2^m , q_3^m]^\top.
$
Since we parameterize rotations as unit quaternions, i.e., $|\textbf{q}^m| = 1$, the direct application of the forward process of the diffusion steps is not feasible as it may generate rotation values outside the $SO(3)$ manifold~\cite{leach2022denoising}.
Following~\cite{leach2022denoising}, we address this limitation by leveraging the diffusion processes on the Lie group $SO(3)$ (more details are available in the Supplementary Material).
As proposed in~\cite{sellan2022breaking}, we also test the effect of an additional Chamfer Distance Loss term. %

\begin{table*}[t]
    \centering
    \begin{tabularx}{\linewidth}{c@{\hskip 0.10in}  X  | r@{\hskip 0.14in}  r@{\hskip 0.14in} r@{\hskip 0.14in} r@{\hskip 0.14in} | r@{\hskip 0.14in}  r@{\hskip 0.14in}  r@{\hskip 0.14in} r@{\hskip 0.14in}}%
    \hlineB{2}
    & & \multicolumn{8}{c}{\textsc{Dataset}}\\
    & \textsc{Method}  & \multicolumn{4}{c|}{\textbf{PuzzleCelebA}} & \multicolumn{4}{c}{\textbf{PuzzleWikiArts}} \\
    & \textbf{} &  \textbf{6x6} & \textbf{8x8} & \textbf{10x10} & \textbf{12x12} & \textbf{6x6} & \textbf{8x8} & \textbf{10x10} & \textbf{12x12}\\
    \hline

    \multirow{3}{*}{\makecell{Optimization \\ Based}}& Gallagher ~\cite{gallagher2012jigsaw} &  {80.21}  &  {55.18}  & 71.19  & 
    69.81  & {71.88}  & {61.63}  & {54.15}  & {44.68}\\
    &Yu \etal \cite{yu2015solving} & 98.63& \underline{94.65} & 98.33 & 93.33& \textbf{94.62}  & \textbf{92.95} & \textbf{90.99} & \textbf{89.88}\\
    &Huroyan \etal \cite{huroyan2020solving} &  98.47 & \textbf{97.45} & 98.65 &  \underline{97.08}&   \underline{92.69} & \underline{91.37} & \underline{89.74} &  \underline{88.28}  \\
    \hline

    \multirow{3}{*}{\makecell{Learning \\ Based}} 
    &\textbf{\mname} - No Diff. &  \underline{99.43} & 79.84 & \underline{99.05} & 91.28 & 73.07 & 54.70 & 22.68 & 18.27 \\
    &\textbf{\mname} - No Equiv. &  96.12 & 71.62 & 91.98 & 64.15 & 25.31 & 14.63 & 8.19 & 4.96 \\
    &\textbf{\mname} &  \textbf{99.51} & 87.66 & \textbf{99.30} & \textbf{97.76} & 90.65 & 72.79 & 63.33 & 53.08\\
    \end{tabularx}
    \centering
    \caption{Results for Jigsaw puzzle solving on \textit{PuzzleCelebA} and \textit{PuzzleWikiArts}}
    \label{tab:puzzle}
    \vspace{-.5cm}
\end{table*}
\begin{table}[t]
    \centering
    \begin{tabularx}{\linewidth}{X | c@{\hskip 0.1in}  c@{\hskip 0.1in} | c@{\hskip 0.1in}  c@{\hskip 0.1in}  }
    \hlineB{2}
    & \multicolumn{4}{c}{\textsc{Dataset}}\\
    \textsc{Method}  & \multicolumn{2}{c|}{\textbf{CelebA}} & \multicolumn{2}{c}{\textbf{WikiArts}} \\
     &  \textbf{6x6} & \textbf{12x12} & \textbf{6x6} & \textbf{12x12} \\
    \hline
    
    \multirow{2}{*}{Gallagher \cite{gallagher2012jigsaw}} &  33.28 & 19.18 & 32.19  & 24.12  \\[-.5ex]
    & \textit{\small{(-46.93)}}  &  \textit{\small{(-50.63)}}& \textit{\small{(-39.69)}}  & \textit{\small{(-20.56)}} \\[.2ex] 

    \multirow{2}{*}{Yu \cite{huroyan2020solving}} &  \underline{33.45} & \underline{21.78} & \underline{32.53}  & \underline{24.65} \\ [-.5ex]
   & \textit{\small{(-66.85)}}  &  \textit{\small{(-72.84)}}& \textit{\small{(-62.09)}}  & \textit{\small{(-65.23)}} \\[.2ex]  
    
   \multirow{2}{*}{Huroyan \cite{yu2015solving}} &  18.18 & 0.09 & 17.14  & 0.08  \\ [-.5ex]
   & \textit{\small{(-80.29)}}  &  \textit{\small{(-88.45)}}& \textit{\small{(-75.55)}}  & \textit{\small{(-80.28)}} \\[.2ex]

   \hdashline
    
    \multirow{2}{*}{\textbf{\mname}} &  \textbf{96.92}  & \textbf{76.49} & \textbf{51.21}  & \textbf{27.09}   \\ [-.5ex]
    & \textit{\small{(-2.59)}} & \textit{\small{(-32.81)}}  & \textit{\small{(-39.44)}}  & \textit{\small{(-25.99)}}\\
    \end{tabularx}
    \centering
    \caption{Results for Jigsaw puzzle solving with 30\% missing pieces on \textit{PuzzleCelebA} and \textit{PuzzleWikiArts}. The percentage variance of the model relative to the result presented in Table~\ref{tab:puzzle} is reported within square brackets.
    }
    \label{tab:puzzle_missing}
    \vspace{-0.5cm}
\end{table}

\paragraph{Results.}
We report in Table~\ref{table:bb_compare} the results of the comparison on BB.
Among the baselines, SE(3)-Equiv, which is the current SOTA, performs best in terms of RMSE(R), and DGL performs best in terms of RMSE(T) and PA.
These baselines trade accuracy in rotation with accuracy in translation, with SE(3)-Equiv performing well in rotation and worst in translation and DGL performing well in translation and badly in rotation. Contrarily, \mname\ outperforms the baselines on all metrics: rotation, translation, and part accuracy, showing the effectiveness of our approach.

When comparing the two variants of our approach, we see the importance of using both an equivariant feature representation and the diffusion process.
Notably, we observe significantly worse results when one of these elements is missing: RMSE(R) drops by $\sim10$ points, RMSE(T) drops by $\sim3$ points, and part accuracy drops from 27\% to %
3\%.

In the following paragraph, we report the results with other variants of our approach.
Figure~\ref{fig:qual_3d} reports a qualitative comparison between \mname\ and the SE(3)-Equiv when reassembling a wine glass and a bottle, both fragmented in four pieces. We observe that SE(3)-Equivariant struggles in dealing with both large and small pieces, and shifts all pieces to the middle point. Contrarily, \mname\ was able to handle big pieces well but struggled with small pieces like the fragments of the glass stem.

\paragraph{Ablation.}
Table~\ref{tab:abl_3d} reports results  %
assessing \textit{i)} the importance of the feature representation, \textit{ii)} the impact of the diffusion design, and \textit{iii)} the benefit of using attention.

We notice that employing invariant and non-equivariant features leads to worse performance. This result highlights the importance of providing the network with inductive biases, specifically rotation-equivariant feature, to solve this task.
In the Diffusion Design section of the table, we compare our implemented forward process for handling rotation in $SO(3)$ with a direct application of Gaussian Noise to the 6D representation (6DOF)~\cite{zhou2019continuity}. This straightforward approach negatively impacts the model's performance across all metrics. 
Following \cite{sellan2022breaking}, we investigate the use of the Chamfer Distance (CD) loss alongside our general losses. We see that by using the CD loss, we improve in Part Accuracy but perform worse in both RMSE for translation and rotation. Nevertheless, the loss in performance is minor, and with or without Chamfer Distance loss, aside from the part accuracy, our method performs best.

Finally, we investigate the impact of the attention mechanism on information propagation. For this purpose, we define the adjacency matrix $A \in \mathbb{R}^{M \times M}$ as an all-ones matrix and, instead of UniMP, we use the Graph Convolutional Network (GCN)~\cite{kipf2016semi}. \mname{} with UniMP consistently outperforms \mname{} with GCN, highlighting the benefit of adopting an attention mechanism.

\subsection{2D Jigsaw Puzzle with Rotated Pieces}\label{sec:2d}

We adopt \mname\ to reassemble visual jigsaw puzzles with translated and rotated pieces. 
In this task, we need to reassemble an image by translating and rotating a collection of image patches. The patches are regularly shaped, non-overlapping, and initialized with a random orientation.

\paragraph{Dataset and Evaluation Setting.}
Following~\cite{talon2022ganzzle}, we test our approach on two datasets of puzzles: PuzzleCelebA~\cite{CelebAMask-HQ} and PuzzleWikiArt~\cite{tan2018improved}.
PuzzleCelebA is a dataset of celebrities' faces, while PuzzleWikiArt contains paintings from various artists in many styles. 
We compare with three optimization-based methods %
for visual puzzle-solving: \textit{i)} Gallagher~\cite{gallagher2012jigsaw}, \textit{ii)} Yu \etal~\cite{yu2015solving}, and \textit{iii)} Huroyan \etal~\cite{huroyan2020solving}. %

We report the results using various numbers of patches, from 36 ($6 \times 6$) to 144 ($12 \times 12$). We present the outcomes using the \emph{direct comparison metric}~\cite{choCVPR10probjigsaw}, where a piece is successfully placed if it is both correctly positioned and rotated. While our solution operates in the continuous domain, the evaluation is conducted in the discrete one of~\cite{gallagher2012jigsaw}. To do so, we first apply to each piece the predicted (continuous) translation and rotation; then, we discretize its pose by
snapping the piece to the closest cell in a $n \times n$ squared lattice, $n = \sqrt{M}$, and its rotation to the 
nearest $\pi/2$ angle. %

\begin{figure}[t]
    
    \includegraphics[width=\linewidth]{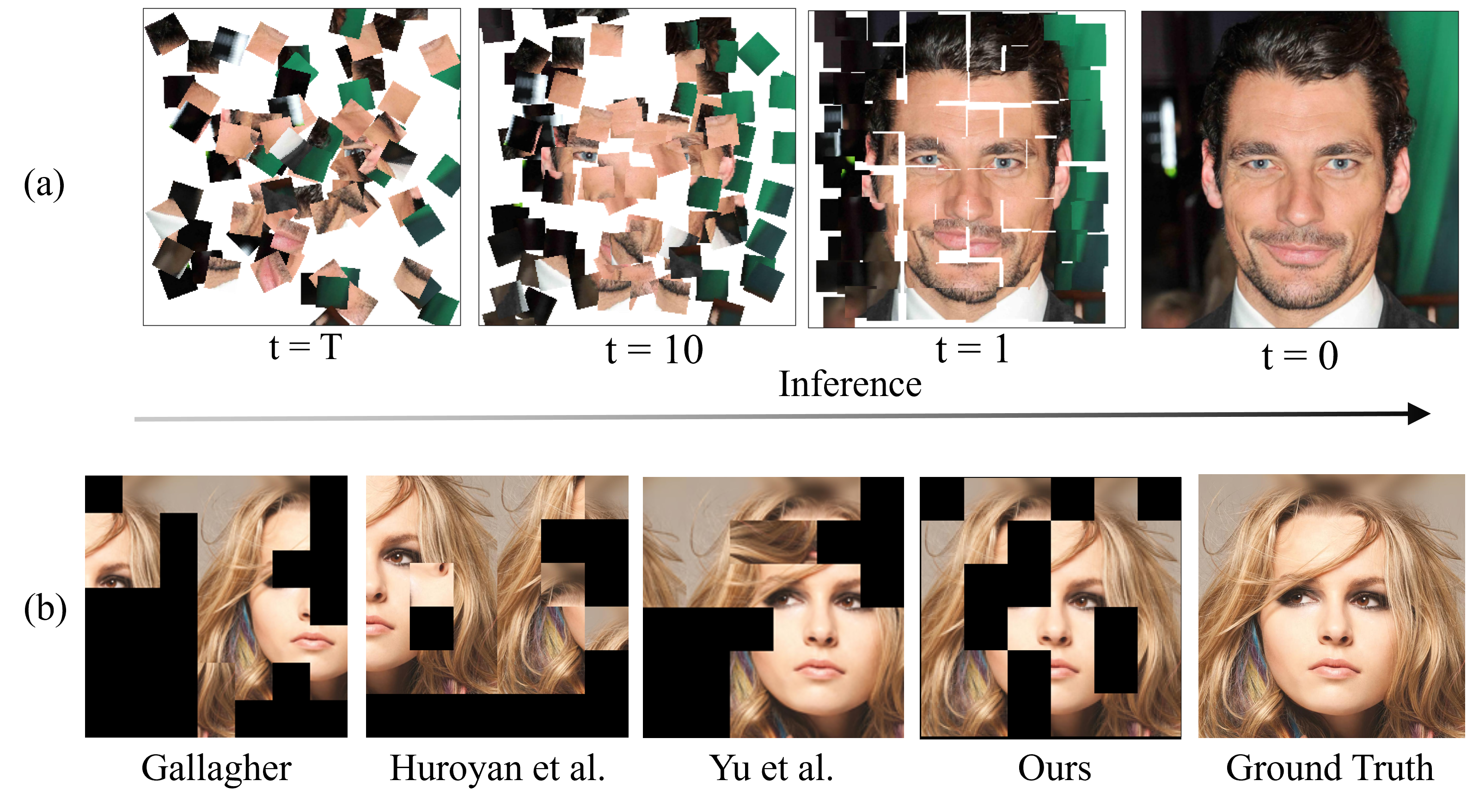}
    \centering
    \caption{    \textbf{(a)} Patch alignment in inference from $t=T$ to $t=0$. \textbf{(b)} Qualitative comparison with $30\%$ missing pieces. 
    \label{fig:qual2d}
    }
    \label{fig:visual_celeba}
    \vspace{-0.5cm}
\end{figure}

\paragraph{Implementation Details.} 
Each piece has to be arranged into a two-dimensional continuous space with boundaries $[-1,1]$.
Following~\cite{gallagher2012jigsaw}, the possible initial rotation of each patch is represented by the set $\{0, \pi/2, \pi, 3\pi/2\}$, which are elements of the cyclic group $\mathcal{Z}_4$~\cite{gallagher2012jigsaw}. 
We parameterized the rotation in 2D as:
        $\textbf{r}^m = [\cos(\theta^m), \sin(\theta^m)]^\top$~\cite{zhou2019continuity} and we optimize $\theta^m$ in the continuous domain.
For the feature extractor, we use a version of ResNet18 that is equivariant~\cite{cohen2016group} to the cyclic group $\mathcal{Z}_4$.

\paragraph{Results.} Table \ref{tab:puzzle} reports results for the visual puzzle reassembly task, with rotated and translated pieces. 
 \mname\ achieves SOTA results in CelebA, improving over the optimization-based method. 
 In Wikiart, the optimization-based approaches \cite{yu2015solving,huroyan2020solving} outperform \mname. An explanation for this gap is that our method relies not only on pure visual appearances but also on the semantic content of the images. CelebA has a very strong semantic structure, containing images of faces in similar poses.
 In contrast, Wikiart has very diverse images with no predefined structure. 
Optimization-based approaches directly match visual content along the borders. This design makes them a strong baseline when all patches are provided, at a cost of time efficiency, as shown in Figure~\ref{fig:time}. 
Section~\ref{sec:scaling} further discusses efficiency as we scale up to larger puzzles.

Relying on semantics also makes \mname's robust when some of the pieces are missing. Testing robustness to missing pieces is common when solving jigsaw puzzles~\cite{gallagher2012jigsaw,talon2022ganzzle}, as it reflects real-world application, e.g., fresco reconstruction~\cite{brown2008system}. We evaluate \mname\ and all the baselines when $30\%$ of the pieces are randomly removed and report results in Table \ref{tab:puzzle_missing}. \mname\ outperforms Huroyan \etal~\cite{yu2015solving}, the second-best model, in both CelebA and Wikiart. Optimization-based methods experience a significant decrease in accuracy on both $6\times 6$ and $12\times 12$ puzzles, while \mname\ retains high performances even in this challenging setting. 
Figure~\ref{fig:qual2d} shows \mname\ solving a CelebA puzzle from randomly shuffled pieces, along with a comparison with all the baselines when $30\%$ of the pieces are missing. 
In the Supplementary Material, we present an ablation on the design choices for 2D Jigsaw puzzle, analogously to the above-mentioned ablation study for 3D object reassembly.

\begin{figure}[t]
    \centering
    \begin{minipage}{1\linewidth}
        \begin{tikzpicture}
        \begin{axis}[
            height=4.5cm,
            width=0.84\linewidth,
            xlabel={Number of elements},
            ylabel={Memory (GB)},
            xmin=0, xmax=900,
            ymin=0, ymax=60,
            ytick={10,  20,  30, 40, 50},
            xmajorgrids=false,
            ymajorgrids=true,
            grid style={line width=.1pt, draw=gray!10},
            major grid style={line width=.2pt,draw=gray!50},
            legend pos=north west,
            axis lines=left,
            axis on top,
            clip=false
        ]
        \addlegendentry{\footnotesize{\mname}}
        \addplot[color=col4, mark=o,line width=1pt] coordinates {
            (36,2.4)
            (144,4.3)
            (400,12.7)
            (676,27)
            (900,50)
        };
        \addlegendentry{\footnotesize{\mname{} w/ sparsity}}
        \addplot[color=col1, mark=square,line width=1pt] coordinates {
            (36,2.3)
            (144,3.9)
            (400,7.5)
            (676,13)
            (900,20)
        };

       \addlegendentry{\footnotesize{NVIDIA RTX 4090}}
       \addplot[color=red, dashed, line width=2pt] coordinates {
            (0,24)
            (900,24)
        };
        \node[anchor=west,text width=1.3cm] at (axis cs:900,35) {\small{2.5x less memory}};
        \addplot[<->,  line width=0.8pt] coordinates {(900,22) (900,48)};
        \end{axis}
          \vspace{-.5cm}
        \end{tikzpicture}
        \caption{\label{fig:memory}GPU memory consumption by total size.}
        \vspace{0.1cm}
    \end{minipage}

    \begin{minipage}{\linewidth}
        \begin{tikzpicture}
        \begin{axis}[
            width=0.84\linewidth,
            height=4.4cm,
            xlabel={Number of elements},
            ylabel={Time (s)},
            ymax=160,
            legend pos=north west,
            ytick={30,  60,  90, 120, 150},
            axis lines=left,
            xmajorgrids=false,
            ymajorgrids=true,
            grid style={line width=.1pt, draw=gray!10},
            major grid style={line width=.2pt,draw=gray!50},
            axis on top,
            clip=false
        ]
        \addlegendentry{\footnotesize{Huroyang\cite{huroyan2020solving}}}
        \addplot[mark=star, col4,line width=1pt] coordinates {
            (36,1)
            (144,8)
            (400,35)
            (625,72)
            (900,152)
        };

      \addlegendentry{\footnotesize{Yu\cite{yu2015solving}}}
       \addplot[mark=triangle, col3,line width=1pt] coordinates {
           (36,0.3)
           (144,3)
           (400,14)
          (625,42)
           (900,92)
       };

             \addlegendentry{\footnotesize{Gallagher\cite{gallagher2012jigsaw}}}
       \addplot[mark=o, col2,line width=1pt] coordinates {
           (36,0.3)
           (144,4.4)
           (400,23.35)
          (625, 34.11)
           (900,55.10)
       };
        
        \addlegendentry{\footnotesize{\mname}}
        \addplot[mark=square, color=col1, line width=1pt] coordinates {
            (36, 1)
            (144,1)
            (400,2)
            (625,3)
            (900,5)
        };
 
        \node[anchor=west,text width=1.3cm] at (axis cs:910,30) {\footnotesize{11x faster}};
        \addplot[<->,  line width=.8pt] coordinates {(890,10) (890,51)};

        \node[anchor=west,text width=1.3cm] at (axis cs:910,70) {\footnotesize{18x faster}};
        \addplot[<->,  line width=.8pt] coordinates {(900,10) (900,85)};

        \node[anchor=west,text width=1.3cm] at (axis cs:910,120) {\footnotesize{30x faster}};
        \addplot[<->,  line width=.8pt] coordinates {(910,10) (910,148)};
        
        \end{axis}
        \end{tikzpicture}
        \caption{\label{fig:time} Time requirement to solve one puzzle for our approach and the optimization methods. 
        }
    \end{minipage}\hfill
\vspace{-.6cm}

\end{figure}

\subsection{Scaling to Larger Graphs} 
\label{sec:scaling}
We investigate \mname\ with Exphander~\cite{shirzad2023exphormer} for higher-dimensional puzzles. %
We explore the effectiveness of %
scaling our method with PuzzleCelebA puzzles up to 900 pieces ($30\times 30$ puzzles).
\mname{} with Expander prunes 80\% of the edges from the complete graph during training and introduces 8 virtual nodes to ensure global connectivity.
Figure~\ref{fig:memory} shows the memory requirements for \mname\ with and without sparsity, executed on standard consumer-grade hardware (NVIDIA GeForce RTX 4090 with 24GB). When the graph has 900 elements, our method with sparsity halves the memory consumption without compromising accuracy. %

Although our method requires much memory, it is significantly faster than memory-intensive optimization methods.
We compare \mname\ with the three optimization-based approaches. %
Figure~\ref{fig:time} reports the time required for the four methods to solve a puzzle based on size. The time required by optimization-based approaches increases exponentially with the number of elements and, consequently, with the number of matches. 
On the other hand, \mname\ reassembles up to 900 elements without scaling in time requirement, e.g., it solves $30 \times 30$ puzzles in 5s with $95\%$ accuracy. This represents a significant improvement over Gallagher, the faster optimization-based solution, which has a run-time of 55s with an accuracy of $58\%$.

\section{Conclusion} \label{sec:conclusion}

In this work, we introduced \bmname, a general framework for tackling reassembly tasks through graph representations and a diffusion model formulation. By framing reassembly as a denoising task, we leverage an Attention-based Graph Neural Network to iteratively refine the pose of each piece through a diffusion process.

Our experimental evaluation showcases the effectiveness of \mname, spanning 3D object reassembly and 2D puzzles with translated and rotated pieces. The results demonstrate SOTA performance in most 2D and 3D scenarios, revealing a common ground between these seemingly disparate tasks. Notably, in the 2D domain, \mname{} exhibits robustness to missing pieces and achieves remarkable efficiency compared to optimization-based methods. In the 3D, our solution obtains SOTA results, maintaining accuracy in translation and rotation, unlike previous solutions. %

\paragraph{Limitations and Future Research.} One of the main limitations of \mname\ is its high memory usage, even by introducing the sparsity mechanism based on the expander graph. Future efforts will focus on mitigating the memory demands and exploring further reassembling scenarios while dealing with data from real-world scans.

\paragraph{Acknowledgments} This work is part of the RePAIR project that has received funding from the European Union’s Horizon 2020 research and innovation program under grant agreement No. 964854 and is supported by the project Future Artificial Intelligence Research (FAIR) – PNRR MUR Cod. PE0000013 - CUP: E63C22001940006.

\bibliographystyle{ieeenat_fullname}
\bibliography{main}

\clearpage

\clearpage
\setcounter{page}{1}
\maketitlesupplementary
\appendix

\section{Experiment Details}

\paragraph{Hardware.} The experiments were conducted on 2 different machines: four NVIDIA Tesla V100 16GB, 380 GB RAM, and 2x Intel(R) Xeon(R) Silver 4210 CPU @ 2.20GHz Sky Lake CPU, and one NVIDIA RTX 4090 GPU, 64 GB RAM, and 12th Gen Intel(R) Core(TM) i9-12900KF CPU @ 3.20GHz CPU.

\paragraph{Model Settings.} We train \mname{} with a learning rate of $10^{-4}$ and Adagrad as the optimization algorithm~\cite{duchi2011adaptive}. During our training process, we set a maximum of 1000 epochs, but we stop the training earlier to prevent unnecessary iterations when the loss no longer decreases.

\section{Equivariant Feature Representation}

As we presented in Section~\ref{sec:features}, one of key point of our proposal lies in its ability to work with element features $\mathbf{h}^m$, which can be extracted by any pre-trained encoders. In particular, we discover the importance to extract rotation-equivariant features.

A function $\phi$ is equivariant to the action of a group $G$ if $\phi(S_g(\cdot)) = S^{'}_g(\phi(\cdot))$ for all $g \in G$, 
where $S_g$ and $S^{'}_g$ are linear representations related to the group element $g$~\cite{serre1977linear}.
This means that applying $\phi$ to the codomain of $S_g(\cdot)$ %
is equivalent to applying $S^{'}_g \in G$ to the codomain of $\phi$. %
In this work, the transformation $S_g$ and $S^{'}_g$ are rotations. As a result, the equivariant function $\phi(\cdot)$, i.e. the backbone, ensures the consistency of the rotational effect irrespective of whether it is applied before or after the function. 
Consequently, \mname{} associates a specific rotation $\textbf{r}^m$ (in the input space) to the features vector $\textbf{h}^m$. %

\section{Diffusion process and Rotation in 3D}\label{appendix:rotation}

We provide a more detailed description of how we introduce Gaussian Noise with 3D rotations. Following~\cite{leach2022denoising}, we use a specific procedure to scale the rotation matrices $f_r({\textbf{r}^m_t})$ by \textit{i)} converting the rotation matrix to values in the Lie algebra $\mathfrak{so}(3)$, \textit{ii)} multiplying them element-wise with $t$-dependent scalars, and \textit{iii)} converting back to a rotation matrix through matrix exponentiation. Analogous to an addition in Euclidean space, the composition of rotations is done through matrix multiplication in $SO(3)$ as:

\begin{equation*}
    \lambda(\gamma_t, \textbf{r}^m_{t}) = \exp(\gamma_t \log(f_r(\textbf{r}^m_t))),
\end{equation*}
where $\lambda(\dots)$ is the geodesic distance flow from \textbf{I}, the identity matrix, to $\textbf{r}^m_{t}$ by an amount $\gamma_t$.

In particular, for the Forward Process, we rewrite Equation~\eqref{eq:forward} %
to inject noise into $\textbf{r}^m_0$:
   
\begin{equation*}
q(\textbf{r}^m_t|\textbf{r}^m_0) = IG_\text{SO(3)}(\lambda(\sqrt{\overline{\alpha_t}}, \textbf{r}^m_0), (1 - \overline \alpha_t)), 
\end{equation*}
where $IG_\text{SO(3)}$ is the isotropic Gaussian distribution (IG) that is compatible with $SO(3)$ rotation directly. The IG distribution is parameterized in an axis-angle form by sampling uniformly an axis and rotation angle $\omega \in [0, \pi]$ as:
\begin{equation*}
    f(\omega) = \frac{1-\cos \omega}{\pi}\sum_{l=0}^{\infty}(2l + 1)e^{-l(l+1)\epsilon^2}\frac{\sin((l+ 0.5) \omega}{\sin(\omega/2)}.    
\end{equation*}
For the Reverse Process, letting ${R}_t = \{\textbf{r}^m_t\}_{m \in [1,\cdots, M]}$ and  ${H} = \{\textbf{h}^m\}_{m \in [1,\cdots, M]}$, we rewrite Equation~\eqref{eq:backward} as follows: %
\begin{equation*}
     \hat{{R}}_{t-1} = \lambda\left( \frac{\sqrt{\overline \alpha_{t-1}}}{\alpha_t}, R_t \right) \lambda
     \left(\frac{1-\overline{\alpha}_{t-1}}{\sqrt{\overline\alpha_t}}, \epsilon_\theta^\text{rot}({R}_t, t, H)\right)^T,
\end{equation*}
where $\epsilon_\theta^\text{rot}({R}_t, t, H)$ is the estimated noise that has to be removed from $R_t$ to recover $\hat{{R}}_{t-1}$.

\section{Additional Ablations}

\begin{table*}[t]
\centering
\begin{tabularx}{\linewidth}{X | P{2cm} P{2cm} P{2cm} | P{2cm} P{2cm} P{2cm}}
\hline
\multicolumn{1}{l|}{Missing} & \multicolumn{3}{c|}{0\%} & \multicolumn{3}{c}{10\%} \\ \hline
\multirow{2}{*}{Method} & RMSE ($R$) $\downarrow$ & RMSE ($T$) $\downarrow$ & PA $\uparrow$ & RMSE ($R$) $\downarrow$ & RMSE ($T$) $\downarrow$ & PA $\uparrow$ \\
 & \multicolumn{1}{c}{degree} & \multicolumn{1}{c}{$\times 10^{-2}$} & \multicolumn{1}{c|}{$\%$} & \multicolumn{1}{c}{degree} & \multicolumn{1}{c}{$\times 10^{-2}$} & \multicolumn{1}{c}{$\%$} \\ \hline
Global & 83.00 & 18.74 & 7.02 & 83.86 & \underline{18.76} & 6.78  \\
DGL & 84.56 & \textbf{18.26}  & \underline{9.72} & 84.74 & 18.98 & \underline{8.42}  \\
LSTM & 88.26 & 19.64 &  4.78 & 88.40 & 19.74 & 4.96  \\
SE(3)-Equiv & \underline{81.82} & \underline{18.50}  &  6.74 & \underline{82.96} & \textbf{18.54}  & 6.58 \\ \hline
\bmname{} & \textbf{80.13} & 19.02 & \textbf{11.61} &  \textbf{80.32} & 19.32 &\textbf{11.20} \\ %
\multicolumn{1}{c}{} \\
\multicolumn{1}{c}{} \\ \hline
\multicolumn{1}{l|}{Missing} & \multicolumn{3}{c|}{20\%} & \multicolumn{3}{c}{30\%} \\ \hline
\multirow{2}{*}{Method} & RMSE ($R$) $\downarrow$ & RMSE ($T$) $\downarrow$ & PA $\uparrow$ & RMSE ($R$) $\downarrow$ & RMSE ($T$) $\downarrow$ & PA $\uparrow$ \\
 & \multicolumn{1}{c}{degree} & \multicolumn{1}{c}{$\times 10^{-2}$} & \multicolumn{1}{c|}{$\%$} & \multicolumn{1}{c}{degree} & \multicolumn{1}{c}{$\times 10^{-2}$} & \multicolumn{1}{c}{$\%$} \\ \hline
Global & 84.20 & \underline{18.86} & 6.66 & 84.76 & \textbf{18.96} & \underline{6.62}  \\
DGL & 85.01 & 19.80 & \underline{7.34} & 85.64 & 20.68 & 6.56  \\
LSTM & 88.72 & 19.90 & 4.88 & 88.96 & 20.01 & 4.36  \\
SE(3)-Equiv & \underline{82.52}  &  \textbf{18.72} & 6.54 & \underline{82.88} & \underline{19.48}  & 6.51  \\ \hline
\bmname{} & \textbf{80.37} & 19.52 & \textbf{10.67} & \textbf{80.46} & 19.84 & \textbf{10.43}

\end{tabularx}%
\caption{Results for \mname\ on BB's objects with 8-20 pieces when 0\%/10\%/20\%/30\% of the pieces are missing pieces. Our approach is robust even in the hardest scenario where 30\% of the pieces are missing.}
\label{tab:3d_missing}
\end{table*}

\begin{table*}[t]
    \centering
    \begin{tabularx}{\linewidth}{c |X | r@{\hskip 0.1in}  r@{\hskip 0.1in} r@{\hskip 0.1in} r@{\hskip 0.1in} |r@{\hskip 0.1in}  r@{\hskip 0.1in}  r@{\hskip 0.1in} r@{\hskip 0.1in}}
    \hlineB{2}
    \textsc{Stage} & \textsc{Changes} & \multicolumn{4}{c|}{\textbf{PuzzleCelebA}} & \multicolumn{4}{c}{\textbf{PuzzleWikiArts}} \\
    &  &  \textbf{6x6} & \textbf{8x8} & \textbf{10x10} & \textbf{12x12} & \textbf{6x6} & \textbf{8x8} & \textbf{10x10} & \textbf{12x12} \\
    \hline
    \multirow{2}{*}{Representation} & Non-Equivariant Enc. &  96.12 & 71.62 & 91.98 & 64.15 & 25.31 & 14.63 & 8.19 & 4.96 \\
    & Invariant Enc. &  22.97 & 20.01 & 16.87 & 13.63   & 7.64 & 4.64 & 2.79 & 1.66  \\
      \hdashline
    \multirow{1}{*}{Diff. Process}    & No Diff. process %
    &  99.43 & 79.84 & 99.05 & 91.28 & 73.07 & 54.70 & 22.68 & 18.27\\
    
    \hdashline
    \multirow{1}{*}{GNN}& Standard GCN~\cite{kipf2016semi} & 85.03 & 54.35 & 71.19& 45.56 & 30.12 & 22.07 &  10.77 & 1.08\\
    \hline
        \textbf{\mname}& Base Implementation (Tab.~{\color{red}3}) &  \textbf{99.51} & \textbf{84.94} & \textbf{99.30} & \textbf{97.76} & \textbf{90.65 }& \textbf{72.79} & \textbf{63.33} & \textbf{53.08}\\
    \end{tabularx}
    \centering
\caption{We conduct an ablation study to evaluate the impact of each component of \mname\ for Jigsaw puzzle solving on \textit{PuzzleCelebA} and \textit{PuzzleWikiArts}. The base implementation corresponds to our proposed approach, as reported in Table {\color{red}3} of the main paper.
    }
    \label{tab:abl_puzzle}
\end{table*}

\paragraph{Missing Fragments in 3D Objects Reassembly} 

We assess the performance of \mname{} and the baselines in scenarios involving missing 3D pieces. We consider a setting where each object is composed of 10 to 20 parts. We test the methods in four different scenarios: \textit{i)} without missing pieces, \textit{ii)} 10\% of missing pieces, \textit{iii)} 20\% missing pieces, and \textit{iv)} 30\% of missing pieces. We do not retrain the models with missing pieces, but instead, we use the same method and weights as in the main paper experiment described in Section~\ref{sec:3d}. 
To account for potential variations in fracture sizes within each object, we report the experiment five times using different seeds. This methodology helps alleviate potential biases introduced by excluding fractures with differing levels of complexity. Mean and standard deviation  for each metric provide an indication of the overall behavior of the compared methods.

Table~\ref{tab:abl_3d} reports the results, demonstrating that in all four scenarios, \mname{} outperforms the baseline in 2 out of 3 metrics. There is a decrease in performance when we increase the number of missing pieces, even if this reduction is minimal.

\paragraph{2D Jigsaw Puzzle.} 

Table~\ref{tab:abl_puzzle} reports further ablation results for the puzzle setting, which we could not include in the main paper due to space constraints.

We assess the benefit of employing rotation-equivariant features, instead of invariant and non-equivariant ones. %
These two last representations lead to worse performance in both datasets. In particular, these differences are more evident with the WikiArt dataset. \mname{} obtains an average improvement of 94.41\% and 82.43\% compared to \mname{} with invariant and non-equivariant features.
This result highlights, one more time, the importance of employing rotation-equivariant features to solve reassembly tasks when rotation is involved. 

We aimed at demonstrating that the adoption of the diffusion process is well-founded and effective. For this reasons, 
we experiment \mname{} wihtout the diffusion process.
The results show that predicting the pose without the diffusion process, i.e., in 1 step, leads to worst performance, which serves as strong justification for the inclusion and use of the diffusion process in our approach.

Finally, we conducted an ablation for  the \textit{GNN architecture} adopted in \mname{}. Specifically, we assess the Graph Convolutional Network (GCN)~\cite{kipf2016semi} against UniMP. The goal is to investigate the impact of the attention mechanism on information propagation. For this purpose, we define the adjacency matrix $A \in \mathcal{R}^{M \times M}$ of the GCN as an all-ones matrix. Tables~\ref{tab:abl_puzzle} reports the results of this comparison in the 2D and 3D scenarios, respectively. \mname{} with the use of UniMP consistently outperforms \mname{} with GCN, showing a remarkable improvement. %
These results highlight the importance of employing a mechanism that can effectively capture relationships among nodes.

\subsection{Effect of Edge Pruning}

Figure~\ref{fig:pruning_results} presents an ablation study on \textit{PuzzleCelebA}, where we vary the pruning rate during training. Increasing the pruning, i.e., reducing the graph size, has a minor effect on the final results.

\begin{figure}
  \centering
  \begin{tikzpicture}
    \begin{axis}[
        width=0.84\linewidth,
        height=4.5cm,
      xlabel={Pruning (\%)},
      ylabel={Accuracy (\%)},
      xtick={0,20,60,80},
      ymin=75,
      ymax=100,
     legend pos=south west,
    xmajorgrids=false,
    ymajorgrids=true,
    grid style={line width=.1pt, draw=gray!10},
    major grid style={line width=.2pt,draw=gray!50},
    axis lines=left,
    axis on top,
    clip=false
      ]
      
      \addplot[mark=square,color=col1, line width=1pt] coordinates {(0,99.51) (20,99.12) (60,99.04) (80,98.57)};
      \addlegendentry{\footnotesize{6$\times$6}}
      
      \addplot[mark=triangle,color=col4, line width=1pt] coordinates {(0,97.76) (20,96.12) (60,94.85) (80,93.51)};
      \addlegendentry{\footnotesize{12$\times$ 12}}
      
    \end{axis}
  \end{tikzpicture}
  \caption{Ablation \dropout\ for Jigsaw puzzle solving %
    on \textit{PuzzleCelebA}. }
  \label{fig:pruning_results}
\end{figure}

\section{Dataset Details}
\paragraph{3D Reassembly Task.}

A 3D reassembly involves aligning fragments of a broken object into its original form, an essential task with applications in artifact preservation, digital heritage archiving, computer vision, robotics, and geometry processing. Despite its practical importance, the field has faced challenges due to the lack of suitable datasets for studying the natural fracture process. Existing datasets, such as PartNet \cite{mo2019partnet}, AutoMate \cite{jones2021automate}, and JoinABLe~\cite{willis2022joinable}, rely on semantic segmentation, failing to represent objects broken under natural, physically realistic conditions. Breaking Bad (BB) \cite{sellan2022breaking} fills this gap by simulating fractures using an algorithm that accounts for an object’s most geometrically natural breaking patterns, thus creating a dataset that more realistically represents the challenges faced in fragments reassembly. \textbf{BB} contains approximately 10,000 meshes sourced from PartNet and Thingi10k. Each mesh includes 80 fractures, resulting in a total of 1,047,400 breakdown patterns. The dataset is divided into three subsets: \textit{everyday}, \textit{artifact}, and \textit{other}. In this work, we focus on the \textit{everyday} subset, as it is the commonly used dataset for evaluation in previous literature \cite{wu2023leveraging}. Qualitative examples can be found in the video attached to this Supplementary Material.

\begin{figure}[t]
\centering
\includegraphics[width=\columnwidth]{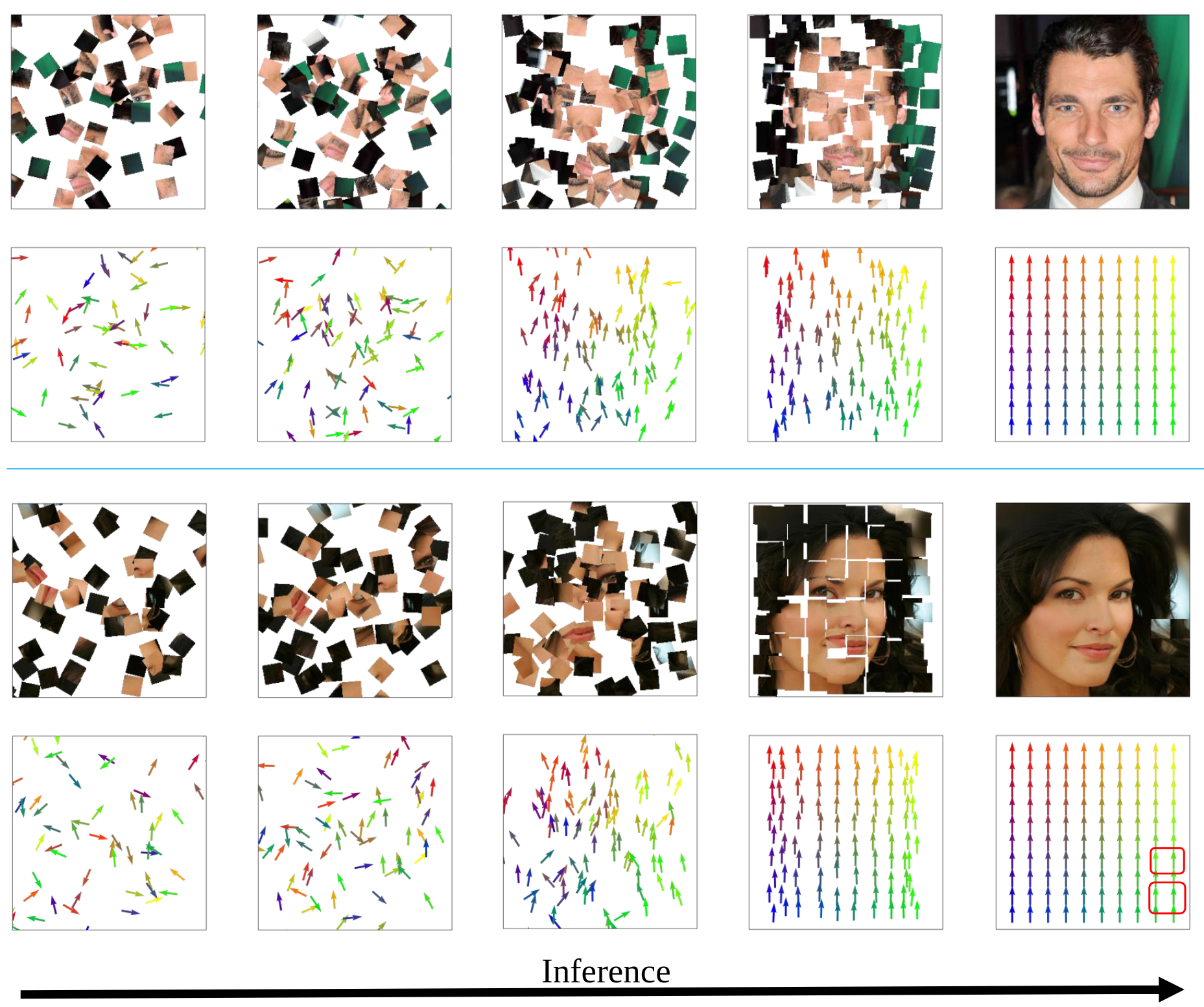}
\caption{\label{fig:puzzle_qual}Qualitative results showing the diffusion process from random to solved puzzle. Each arrows correspond to one piece of the puzzle and its orientation indicate the orientation of the piece.}
\end{figure}

\paragraph{2D Reassembly Task.}
In this task, we evaluated \mname{} on two datasets: PuzzleCelebA and PuzzleWikiArts. Figure~\ref{fig:puzzle_qual} shows some examples of inputs and reconstructions. More examples can be found in the video attached as Supplementary Material.
\begin{itemize}[noitemsep,nolistsep,leftmargin=*]
    \item \textit{PuzzleCelebA} is based on CelebA-HQ~\cite{CelebAMask-HQ} which contains 30K images of celebrities in High Definition (HD). Despite its superficial simplicity, this dataset poses significant challenges for puzzle-solving algorithms due to the inherent symmetry in human faces and often indistinct backgrounds. The dataset is divided in 80-20\% train-test split, with 6,000 test puzzle permutations and randomly rotated patches.
    \item \textit{PuzzleWikiArts} is based on WikiArts~\cite{artgan2018}, and contains 63K images of paintings in HD. This dataset is particularly challenging due to very different content, artistic styles, and intricate patterns, which test the limits of puzzle-solving algorithms. The dataset is split into an 80-20\% train-test ratio, resulting in 50k training images and 13k test puzzles across various grid sizes. It represents a more challenging dataset for puzzle solving as the paintings do not have a common pattern as in PuzzleCelebA (i.e. portraits).
\end{itemize}

\begin{table*}[th]
    \centering
    \begin{tabularx}{\linewidth}{X  r@{\hskip 0.3in}  r@{\hskip 0.3in} r@{\hskip 0.3in} r@{\hskip 0.3in} r@{\hskip 0.3in}  r@{\hskip 0.3in}  r@{\hskip 0.3in} r@{\hskip 0.1in}}
    \hlineB{1}
    \textsc{Method w/ \% degree}  & \multicolumn{8}{c}{\textbf{PuzzleCelebA}} \\
    \textbf{} & \textbf{6x6} & \textbf{8x8} & \textbf{10x10} & \textbf{12x12}  & \textbf{14x14}   & \textbf{16x16}   & \textbf{18x18}   & \textbf{20x20}  \\
    \hline

    \textit{Degree $20\%$ }& \multicolumn{7}{c}{} \\

    Classical dropout & 91.60 & 57.08 & 82.32 & 50.18 & 74.43 & 25.40& 61.45 & 28.35 \\
    \dropout  & \textbf{92.37} &\textbf{ 59.45 }& \textbf{87.67 }&\textbf{ 54.07}& \textbf{83.11 }& \textbf{31.07} &\textbf{ 73.88 }& \textbf{32.97} \\
    \hdashline
    \textit{Degree $60\%$ }& \multicolumn{7}{c}{} \\
    Classical dropout  & \textbf{99.17 }& 72.93 &\textbf{ 98.56} & 94.07 & 98.53& 46.48 &\textbf{ 98.35} & 92.51\\
    \dropout & 99.04 & \textbf{73.91} & 98.43 & \textbf{94.35 }& \textbf{98.70} & \textbf{48.26} & 97.75 & \textbf{93.29} \\
    \hdashline
    \textit{Degree $80\%$} & \multicolumn{7}{c}{} \\
    Classical dropout & 99.15 & 76.45 & 98.75 & 95.87 & 98.58 & 51.51 & 98.14 & 94.93 \\
    
     \dropout& 99.15 & 78.18 & 98.77 & 95.71 & 98.69 & 52.28 & 97.80 & 94.34\\
    \end{tabularx}
    \centering
    \caption{Ablation \dropout\ for Jigsaw puzzle solving %
    on \textit{PuzzleCelebA}. }
    \label{tab:abl_dropout}
\end{table*}

\end{document}